\documentclass[10pt,twocolumn,letterpaper]{article}
\makeatletter
\renewcommand{\paragraph}{%
  \@startsection{paragraph}{4}%
  {\z@}{1.25ex \@plus 1ex \@minus .2ex}{-1em}%
  {\normalfont\normalsize\bfseries}%
}
\makeatother

\usepackage{cvpr}
\usepackage{times}
\usepackage{epsfig}
\usepackage{graphicx}
\usepackage{amsmath}
\usepackage{amssymb}
\usepackage{enumitem}
\usepackage{color}
\usepackage{booktabs}
\usepackage{multirow}
\usepackage{colortbl}
\usepackage{mdwlist}

\setlist{nolistsep}

\newcommand{\abr}[1]{\textsc{#1}}

\newcommand{\ignore}[1]{}

\usepackage[pagebackref=true,breaklinks=true,letterpaper=true,colorlinks,bookmarks=false]{hyperref}

\newcommand{\comics}[0]{{\bf \textsc{comics}}}

\cvprfinalcopy 

\def\cvprPaperID{3408} 

\ifcvprfinal\fi
\begin{document}


\title{The Amazing Mysteries of the Gutter: \\Drawing Inferences Between Panels in Comic Book Narratives}

\author{Mohit Iyyer\thanks{\hspace{.06in}Authors contributed equally}\hspace{0.06in}$^{1}$ \hspace{.1in}Varun Manjunatha\footnotemark[1]\hspace{0.06in}$^{1}$ \hspace{.1in}Anupam Guha$^1$ \hspace{.1in}Yogarshi Vyas$^1$\\
Jordan Boyd-Graber$^2$ \hspace{.1in}Hal Daum\'e III$^1$ \hspace{.1in}Larry Davis$^1$\\
\small{$^1$University of Maryland, College Park\hspace{.1in} $^2$University of Colorado, Boulder}\\
{\tt\small \{miyyer,varunm,aguha,yogarshi,hal,lsd\}@umiacs.umd.edu\hspace{.1in} jordan.boyd.graber@colorado.edu}
}

\maketitle

\begin{abstract}

Visual narrative is often a combination of explicit information
and judicious omissions, relying on the viewer to supply
missing details.  In comics, most movements in time and space are
hidden in the ``gutters'' between panels. To follow the story, readers
logically connect panels together by inferring unseen actions through
a process called ``\textbf{closure}''. While computers can now describe what is explicitly depicted in \underline{natural} images, in this paper we examine
whether they can understand the closure-driven narratives conveyed by
\underline{stylized} artwork and dialogue in comic book panels. We construct a
dataset, \comics, that consists of over 1.2 million panels (120 GB)
paired with automatic textbox transcriptions. An in-depth analysis of
\comics\ demonstrates that neither text nor image alone can tell a comic
book story, so a computer must understand both modalities
to keep up with the plot. We introduce three cloze-style tasks that
ask models to predict narrative and character-centric aspects of a
panel given \emph{n} preceding panels as context. Various deep neural
architectures underperform human baselines on these tasks, suggesting
that \comics\ contains fundamental challenges for both vision and language.

\end{abstract}

\section{Introduction}
\label{sec:intro}

\begin{figure}[t]
    \includegraphics[width=1.0\linewidth]
                   {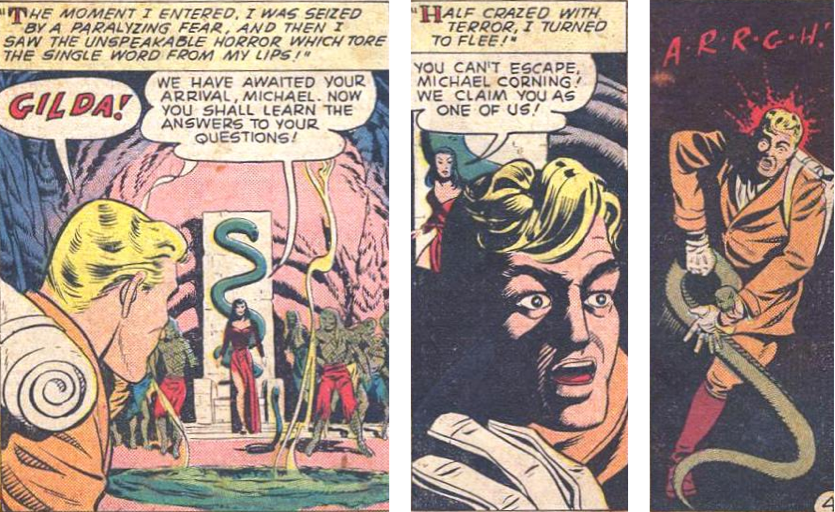}
    \caption{Where did the snake in the last panel come from? Why is
      it biting the man? Is the man in the second panel the same as
      the man in the first panel? To answer these questions, readers form a larger meaning out of the narration boxes, speech bubbles,
      and artwork by applying closure across panels.}
    	\label{fig:panel_transition}
\end{figure}

Comics are fragmented scenes forged into full-fledged stories by the
imagination of their readers. A comics creator can condense anything
from a centuries-long intergalactic war to an ordinary family dinner
into a single panel. But it is what the creator \emph{hides} from
their pages that makes comics truly interesting: the unspoken
conversations and unseen actions that lurk in the spaces (or gutters)
between adjacent panels. For example, the dialogue in
Figure~\ref{fig:panel_transition} suggests that between the second and
third panels, Gilda commands her snakes to chase after a frightened
Michael in some sort of strange cult initiation. Through a process
called \emph{closure}~\cite{mccloud1994understanding}, which involves
(1) understanding individual panels and (2) making connective
inferences across panels, readers form coherent storylines from
seemingly disparate panels such as these. In this paper, we study whether computers can do the same by collecting a dataset of comic books (\comics) and designing several tasks that require closure to solve.

\begin{figure}[t]
	\centering
    \includegraphics[width=1.0\linewidth]
                   {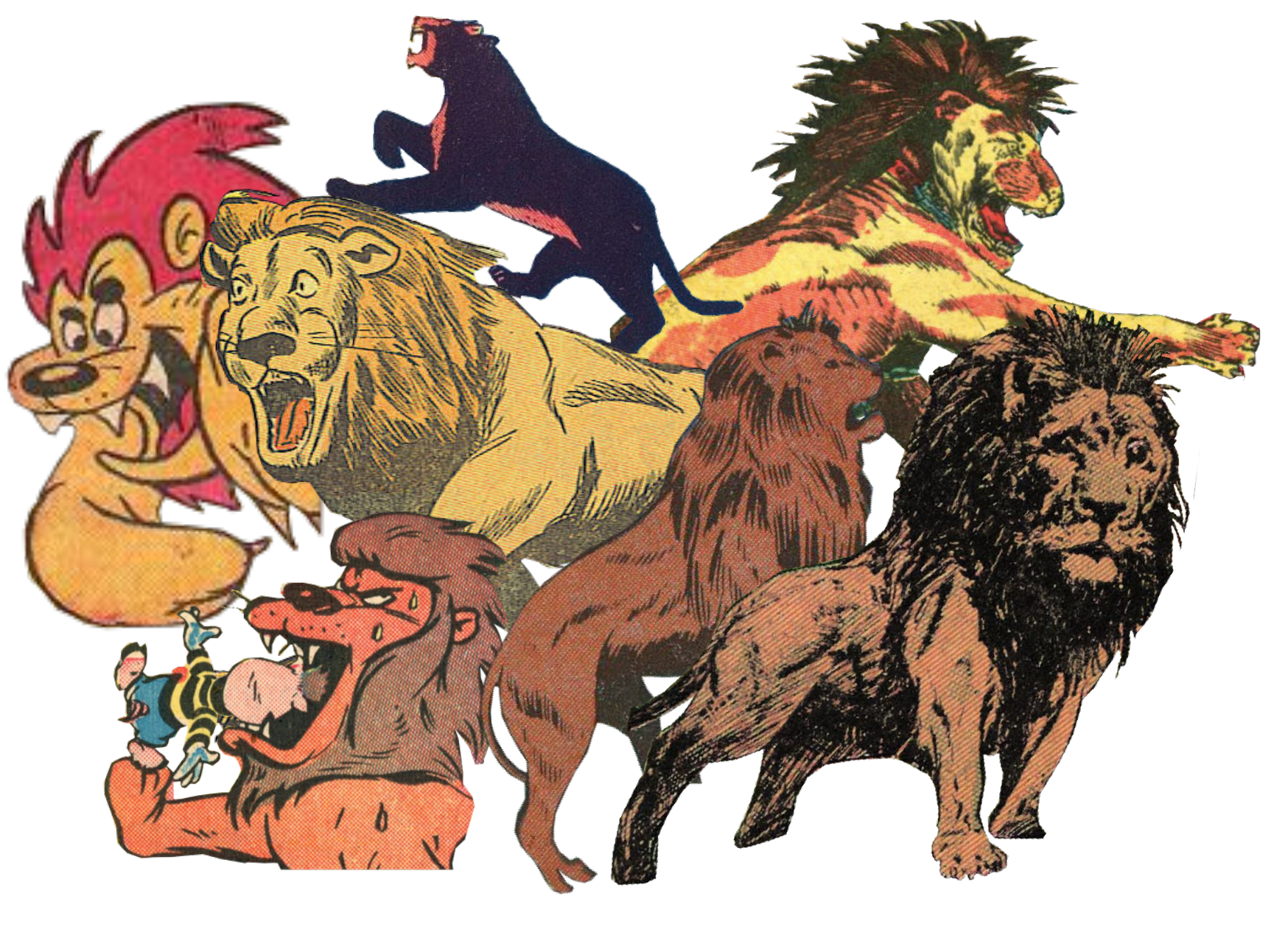}
    \caption{Different artistic renderings of lions taken from the
      \comics\ dataset. The left-facing lions are more cartoonish (and
      humorous) than the ones facing right, which come from action and
      adventure comics that rely on realism to provide thrills.}
    \label{fig:lion_spectrum}
\end{figure}

Section~\ref{sec:dataset} describes how we create
\comics,\footnote{Data, code, and annotations to be made available
  after blind review.} which contains ${\sim}1.2$ million panels drawn
from almost 4,000 publicly-available comic books published during the
``Golden Age'' of American comics (1938--1954).  \comics\ is
challenging in both style and content compared to natural images
(e.g., photographs), which are the focus of most existing datasets and
methods~\cite{krizhevsky2012imagenet,Xu2015show,xiong2016dynamic}. Much
like painters, comic artists can render a single object or concept in
multiple artistic styles to evoke different emotional responses from
the reader. For example, the lions in Figure~\ref{fig:lion_spectrum}
are drawn with varying degrees of realism: the more cartoonish lions,
from humorous comics, take on human expressions (e.g., surprise,
nastiness), while those from adventure comics are more photorealistic.

Comics are not just visual: creators push their
stories forward through text---speech balloons, thought clouds, and
narrative boxes---which we identify and transcribe using optical
character recognition (\abr{ocr}). Together, text and image are often
intricately woven together to tell a story that neither could tell on
its own (Section~\ref{sec:data_analysis}). To understand a story,
readers must connect dialogue and narration to characters and
environments; furthermore, the text must be read in the proper order,
as panels often depict long scenes rather than individual
moments~\cite{cohn2010limits}. Text plays a much larger role in
\comics\ than it does for existing datasets of visual
stories~\cite{Huang2016naacl}.

To test machines' ability to
perform closure, we present three novel cloze-style tasks in
Section~\ref{sec:tasks} that require a deep understanding of narrative
and character to solve. In Section~\ref{sec:models}, we design four neural architectures to examine the impact of multimodality and contextual understanding via closure. 
All of these models
perform significantly worse than humans on our tasks; we conclude
with an error analysis (Section~\ref{sec:discussion}) that suggests
future avenues for improvement.

\section{Creating a dataset of comic books}
\label{sec:dataset}

Comics, defined by cartoonist Will Eisner as \emph{sequential
  art}~\cite{eisner1990comics}, tell their stories in sequences of
\emph{panels}, or single frames that can contain both images and
text. Existing comics
datasets~\cite{guerin2013ebdtheque,matsui2015sketch} are too small to
train data-hungry machine learning models for narrative understanding;
additionally, they lack diversity in visual style and genres. Thus, we build our own dataset,
\comics, by (1) downloading comics in the public domain, (2)
segmenting each page into panels, (3) extracting textbox locations
from panels, and (4) running \abr{ocr} on textboxes and post-processing the
output. Table~\ref{table:size_statistics} summarizes the contents of
\comics. The rest of this section describes each step of our data
creation pipeline.

\subsection{Where do our comics come from?}

\begin{table}
\begin{center}
\begin{tabular}{lr}
\toprule
\# Books & 3,948 \\
\# Pages & 198,657 \\
\# Panels & 1,229,664 \\
\# Textboxes & 2,498,657 \\
\midrule
Text cloze instances & 89,412\\ 
Visual cloze instances & 587,797\\ 
Char. coherence instances & 72,313\\ 
\bottomrule
\end{tabular}
\end{center}
\caption{Statistics describing dataset size (top) and the number of total instances for each of our three tasks (bottom).}
\label{table:size_statistics}
\end{table}

The ``Golden Age of Comics'' began during America's Great Depression
and lasted through World War II, ending in the mid-1950s with the
passage of strict censorship regulations. In contrast to the long,
world-building story arcs popular in later eras, Golden Age comics
tend to be small and self-contained; a single book usually contains
multiple different stories sharing a common theme (e.g., crime or
mystery). While the best-selling Golden Age comics tell of American
superheroes triumphing over German and Japanese villains, a variety of
other genres (such as romance, humor, and horror) also enjoyed popularity~\cite{goulart2004comic}. The Digital Comics Museum
(\abr{dcm})\footnote{\url{http://digitalcomicmuseum.com/}} hosts user-uploaded scans of many comics by lesser-known Golden
Age publishers that are now in the public domain due to copyright
expiration. To avoid off-square images and missing pages, as the scans
vary in resolution and quality, we download the 4,000
highest-rated comic books from \abr{dcm}.\footnote{Some of the panels in \comics\ contain offensive caricatures and opinions reflective of that period in American history.}

\subsection{Breaking comics into their basic elements}
The \abr{dcm} comics are distributed as compressed archives of \abr{jpeg} page
scans. To analyze closure, which occurs from panel-to-panel, we first extract panels from the page images. Next, we extract
textboxes from the panels, as both location and content of textboxes
are important for character and narrative understanding.

\paragraph{Panel segmentation: } 
Previous work on panel segmentation uses
heuristics~\cite{li2014automatic} or algorithms such as density
gradients and recursive
cuts~\cite{tanaka2007layout,pang2014robust,rigaud2015knowledge} that
rely on pages with uniformly white backgrounds and clean
gutters. Unfortunately, scanned images of eighty-year old comics do
not particularly adhere to these standards; furthermore, many \abr{dcm}
comics have non-standard panel layouts and/or textboxes that extend
across gutters to multiple panels.

After our attempts to use existing panel segmentation software failed, we turned to deep learning. We annotate
500 randomly-selected pages from our dataset with rectangular bounding
boxes for panels. Each bounding box encloses both the panel artwork
and the textboxes within the panel; in cases where a
textbox spans multiple panels, we necessarily also include portions of
the neighboring panel. After annotation, we train a region-based
convolutional neural network to automatically detect panels. In
particular, we use Faster \abr{r-cnn}~\cite{renNIPS15fasterrcnn}
initialized with a pretrained \texttt{VGG\_CNN\_M\_1024}
model~\cite{chatfield2014return} and alternatingly optimize the region
proposal network and the detection network. In Western comics, panels
are usually read left-to-right, top-to-bottom, so we also have to
properly order all of the panels within a page after extraction. We compute the midpoint of each panel and sort them using Morton
order~\cite{morton1966computer}, which gives incorrect orderings only
for rare and complicated panel layouts.

\paragraph{Textbox segmentation:} 
Since we are particularly interested in modeling the interplay between
text and artwork, we need to also convert the text in each panel to a
machine-readable format.\footnote{Alternatively, modules for text
  spotting and recognition~\cite{jaderberg2016reading} could be built
  into architectures for our downstream tasks, but since comic
  dialogues can be quite lengthy, these modules would
  likely perform poorly.} As with panel segmentation, existing comic
textbox detection algorithms~\cite{ho2012panel,rigaud2013active} could
not accurately localize textboxes for our data. Thus, we resort again
to Faster \abr{r-cnn}: we annotate 1,500 panels for
textboxes,\footnote{We make a distinction between \emph{narration} and
  \emph{dialogue}; the former usually occurs in strictly rectangular
  boxes at the top of each panel and contains text describing or
  introducing a new scene, while the latter is usually found in speech
  balloons or thought clouds.} train a Faster-\abr{r-cnn}, and sort the
extracted textboxes within each panel using Morton order.

\subsection{OCR}
The final step of our data creation pipeline is applying \abr{ocr} to
the extracted textbox images. We unsuccessfully experimented with two
trainable open-source \abr{ocr} systems,
Tesseract~\cite{smith2007overview} and
Ocular~\cite{berg2013unsupervised}, as well as Abbyy's consumer-grade
FineReader.\footnote{\url{http://www.abbyy.com}} The ineffectiveness
of these systems is likely due to the considerable variation in comic
fonts as well as domain mismatches with pretrained language models
(comics text is always capitalized, and dialogue phenomena such as
dialects may not be adequately represented in training data). Google's
Cloud Vision \abr{ocr}\footnote{\url{http://cloud.google.com/vision}}
performs much better on comics than any other system we tried. While
it sometimes struggles to detect short words or punctuation marks, the
quality of the transcriptions is good considering the image domain and
quality. We use the Cloud Vision API to run \abr{ocr} on all 2.5
million textboxes for a cost of \$3,000. We post-process the
transcriptions by removing systematic spelling errors (e.g., failing
to recognize the first letter of a word). Finally, each book in our
dataset contains three or four full-page product advertisements; since
they are irrelevant for our purposes, we train a classifier on the
transcriptions to remove them.\footnote{See supplementary material for
  specifics about our post-processing.}

\section{Data Analysis}
\label{sec:data_analysis}

\begin{figure*}[t]
	\centering
	\includegraphics[width=1.0\linewidth]
	{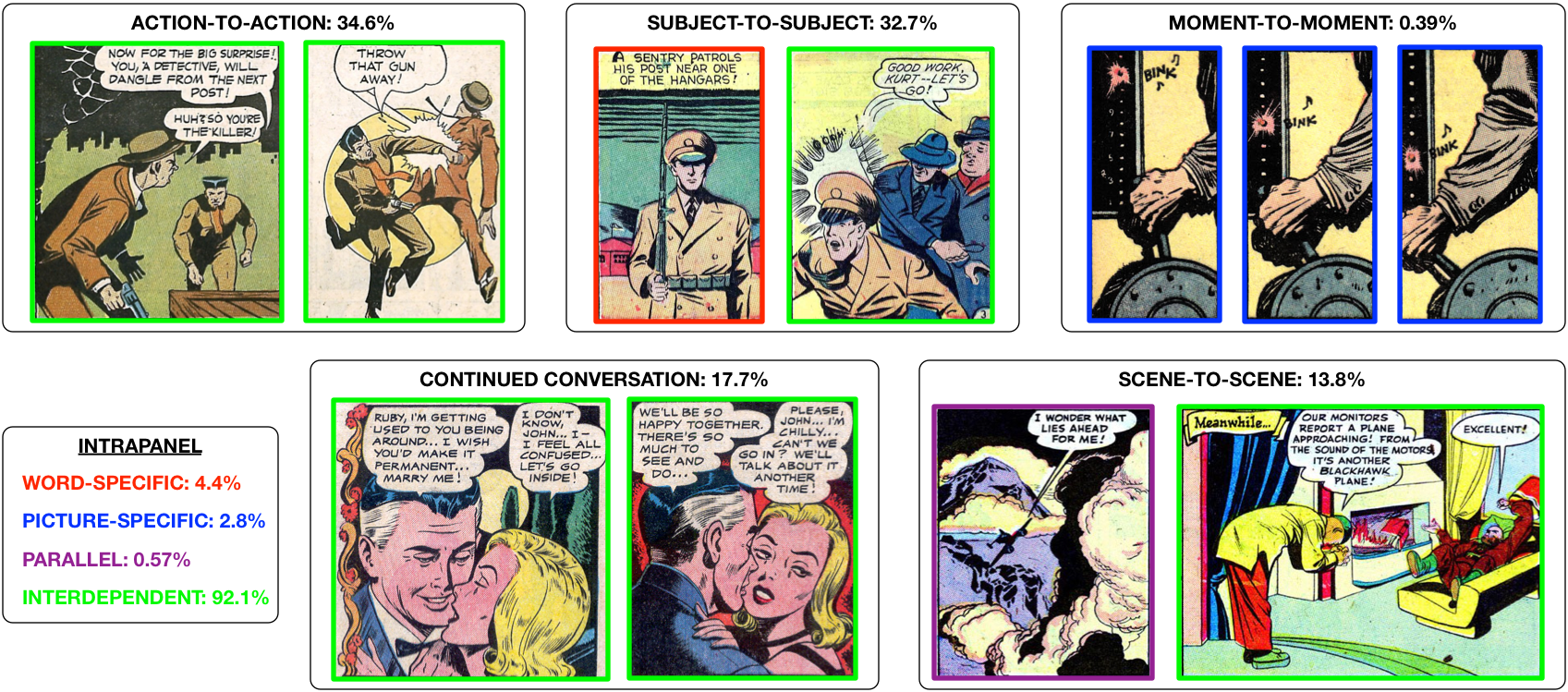}
	\caption{Five example panel sequences from \comics, one for
          each type of interpanel transition. Individual panel borders
          are color-coded to match their intrapanel categories (legend
          in bottom-left). Moment-to-moment transitions unfold like
          frames in a movie, while scene-to-scene transitions are
          loosely strung together by narrative boxes.  Percentages are
          the relative prevalance of the transition or panel type in an annotated subset
          of \comics.}
      	\label{fig:panel_analysis}
\end{figure*}

In this section, we explore what makes understanding narratives in \comics\ difficult, focusing specifically on
\emph{intrapanel} behavior (how images and text interact within a
panel) and \emph{interpanel} transitions (how the narrative advances
from one panel to the next).  We characterize panels and transitions using a
modified version of the annotation scheme in Scott McCloud's
``Understanding Comics''~\cite{mccloud1994understanding}. Over 90\% of
panels rely on both text and image to convey information, as opposed
to just using a single modality. Closure is also important: to
understand most transitions between panels, readers must make complex
inferences that often require common sense (e.g., connecting jumps in space and/or time, recognizing
when new characters have been introduced to an existing scene). We
conclude that any model trained to understand narrative flow in
\comics\ will have to effectively tie together multimodal inputs
through closure.

To perform our analysis, we manually annotate 250 randomly-selected
pairs of consecutive panels from \comics. Each panel of a pair is
annotated for intrapanel behavior, while an interpanel annotation is
assigned to the transition between the panels. Two annotators
independently categorize each pair, and a third annotator makes the
final decision when they disagree. We use four intrapanel categories
(definitions from McCloud, percentages from our annotations):
\begin{enumerate*}
	\item \textbf{Word-specific, 4.4\%}: The pictures illustrate, but
          do not significantly add to a largely complete text.
	\item \textbf{Picture-specific, 2.8\%}: The words do little more than
          add a soundtrack to a visually-told sequence.
	\item \textbf{Parallel, 0.6\%}: Words and pictures seem to follow
          very different courses without intersecting.
	\item \textbf{Interdependent, 92.1\%}: Words and pictures go
          hand-in-hand to convey an idea that neither could convey
          alone.
\end{enumerate*}

We group interpanel transitions into five categories:
\begin{enumerate*}
		\item \textbf{Moment-to-moment, 0.4\%}: Almost no time passes between
		panels, much like adjacent frames in a video.
		\item \textbf{Action-to-action, 34.6\%}: The same subjects progress
		through an action within the same scene.
		\item \textbf{Subject-to-subject, 32.7\%}: New subjects are introduced while staying within the
		same scene or idea.
		\item \textbf{Scene-to-scene, 13.8\%}: Significant changes in time or
		space between the two panels.
		\item \textbf{Continued conversation, 17.7\%}: Subjects continue a conversation across panels without any other changes.
\end{enumerate*}

The two annotators agree on 96\% of the intrapanel annotations
(Cohen's $\kappa = 0.657$), which is unsurprising because almost every
panel is interdependent. The interpanel task is significantly harder:
agreement is only 68\% (Cohen's $\kappa = 0.605$). Panel transitions
are more diverse, as all types except moment-to-moment are relatively
common (Figure~\ref{fig:panel_analysis}); interestingly,
moment-to-moment transitions require the least amount of closure as
there is almost no change in time or space between the panels. Multiple transition types may occur in the same panel, such as simultaneous changes in subjects and actions, which also contributes to the lower interpanel agreement.
\section{Tasks that test closure}
\label{sec:tasks}

\begin{figure*}[t]
	\centering
	\includegraphics[width=1.0\linewidth]
	{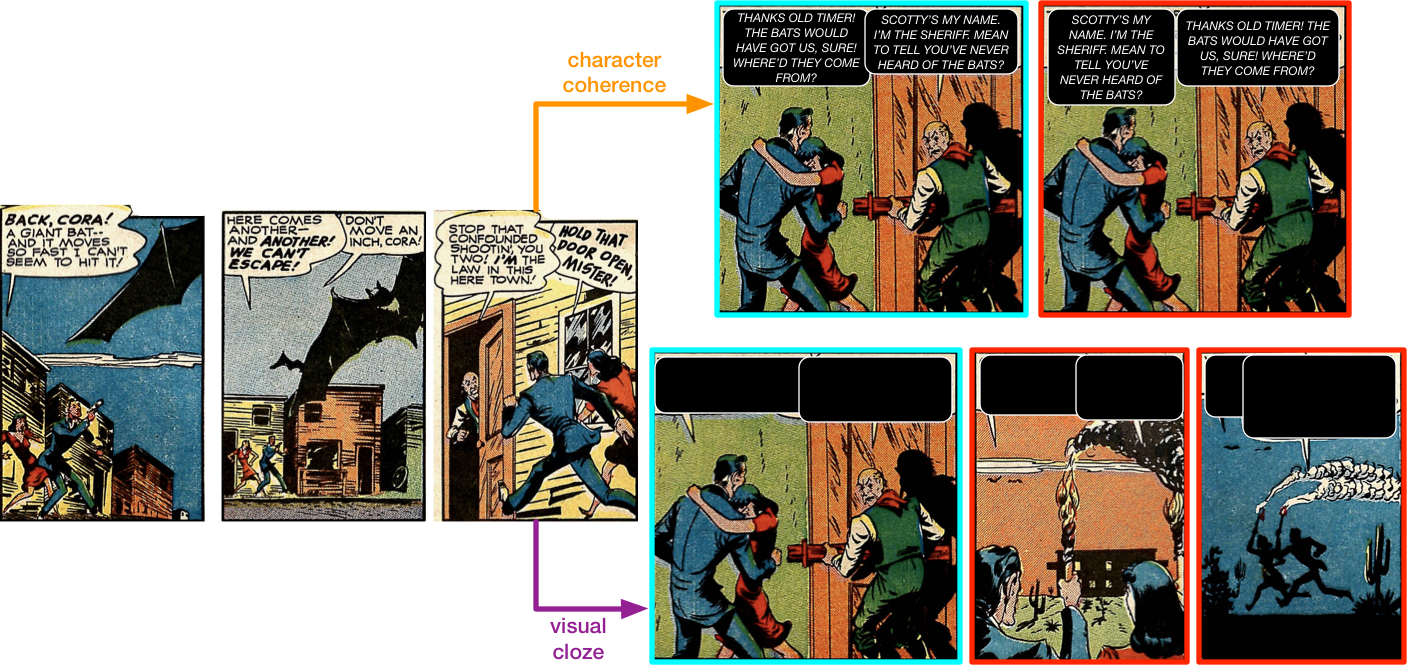}
	\caption{In the
          character coherence task (top), a model must order the dialogues in the final
          panel, while visual cloze (bottom) requires choosing the image of
          the panel that follows the given context. For visualization purposes, we show the original context panels; during model training and evaluation, textboxes are blacked out in every panel.}
       	\label{fig:two_tasks}
\end{figure*}

To explore closure in \comics, we design three novel tasks (\emph{text
  cloze}, \emph{visual cloze}, and \emph{character coherence}) that
test a model's ability to understand narratives and characters given a
few panels of context. As shown in the previous section's analysis, a
high percentage of panel transitions require non-trivial inferences
from the reader; to successfully solve our proposed tasks, a model
must be able to make the same kinds of connections. 

While their objectives are different, all three tasks follow the same
format: given preceding panels $p_{i-1}, p_{i-2},\dots, p_{i-n}$ as
context, a model is asked to predict some aspect of panel $p_i$. While
previous work on visual storytelling focuses on \emph{generating} text
given some context~\cite{huang2016visual}, the dialogue-heavy text in
\comics\ makes evaluation difficult (e.g., dialects, grammatical
variations, many rare words). We want our evaluations to focus
specifically on closure, not generated text quality, so we instead use
a cloze-style framework~\cite{taylor1953cloze}: given $c$
candidates---with a single correct option---models must use the
context panels to rank the correct candidate higher than the
others. The rest of this section describes each of the three tasks in
detail; Table~\ref{table:size_statistics} provides the total instances
of each task with the number of context panels $n=3$.

\paragraph{Text Cloze:}
In the \emph{text cloze} task, we ask the model to predict what text
out of a set of candidates belongs in a particular textbox, given both
context panels (text and image) as well as the current panel
image. While initially we did not put any constraints on the task
design, we quickly noticed two major issues. First, since the panel
images include textboxes, any model trained on this task could
in principle learn to crudely imitate \abr{ocr} by matching text candidates
to the actual image of the text. To solve this problem, we ``black out'' the rectangle
given by the bounding boxes for each textbox in a panel (see
Figure~\ref{fig:two_tasks}).\footnote{To reduce the chance of models trivially correlating candidate length to textbox size, we remove very short and very long candidates.} Second, panels often have multiple
textboxes (e.g., conversations between characters); to focus
on interpanel transitions rather than intrapanel complexity, we
restrict $p_i$ to panels that contain only a single textbox. Thus,
nothing from the current panel matters other than the artwork; the
majority of the predictive information comes from previous panels.

\paragraph{Visual Cloze:}
We know from Section~\ref{sec:data_analysis} that in most cases, text
and image work interdependently to tell a story. In the \emph{visual
  cloze} task, we follow the same set-up as in \emph{text cloze}, but
our candidates are images instead of text. A key difference is that models are not given text from the final panel; in \emph{text cloze}, models are allowed to look at the final panel's artwork. This design is motivated by eyetracking studies in single-panel cartoons, which show that readers look at artwork before reading the text~\cite{carroll1992visual}, although atypical font style and text length can invert this order~\cite{cohn16}. 

\paragraph{Character Coherence:}

While the previous two tasks focus mainly on narrative structure, our
third task attempts to isolate character understanding through a
re-ordering task. Given a jumbled set of text from the textboxes in
panel $p_i$, a model must learn to match each candidate to its
corresponding textbox. We restrict this task to panels that contain
exactly two dialogue boxes (narration boxes are excluded to focus the
task on characters). While it is often easy to order the text based on
the language alone (e.g., ``how's it going'' always
comes before ``fine, how about you?''), many cases require inferring
which character is likely to utter a particular bit of dialogue based
on both their previous utterances and their appearance (e.g.,
Figure~\ref{fig:two_tasks}, top).

\subsection{Task Difficulty}
For \emph{text cloze} and \emph{visual cloze}, we have two difficulty settings that vary in how cloze candidates
are chosen. In the \emph{easy} setting, we sample textboxes (or panel images) from the
entire \comics\ dataset at random. Most incorrect candidates in the
easy setting have no relation to the provided context, as they come
from completely different books and genres. This setting is thus easier for models to
``cheat'' on by relying on stylistic indicators instead of contextual information. With that said, the task
is still non-trivial; for example, many bits of short dialogue can be applicable
in a variety of scenarios. In the
\emph{hard} case, the candidates come from nearby pages, so
models must rely on the context to perform well. For \emph{text cloze}, all candidates are
likely to mention the same character names and entities, while color schemes and textures become much less distinguishing for \emph{visual cloze}.
\section{Models \& Experiments}
\label{sec:models}

\begin{figure*}[t]
	\centering
	\includegraphics[width=0.9\linewidth]
	{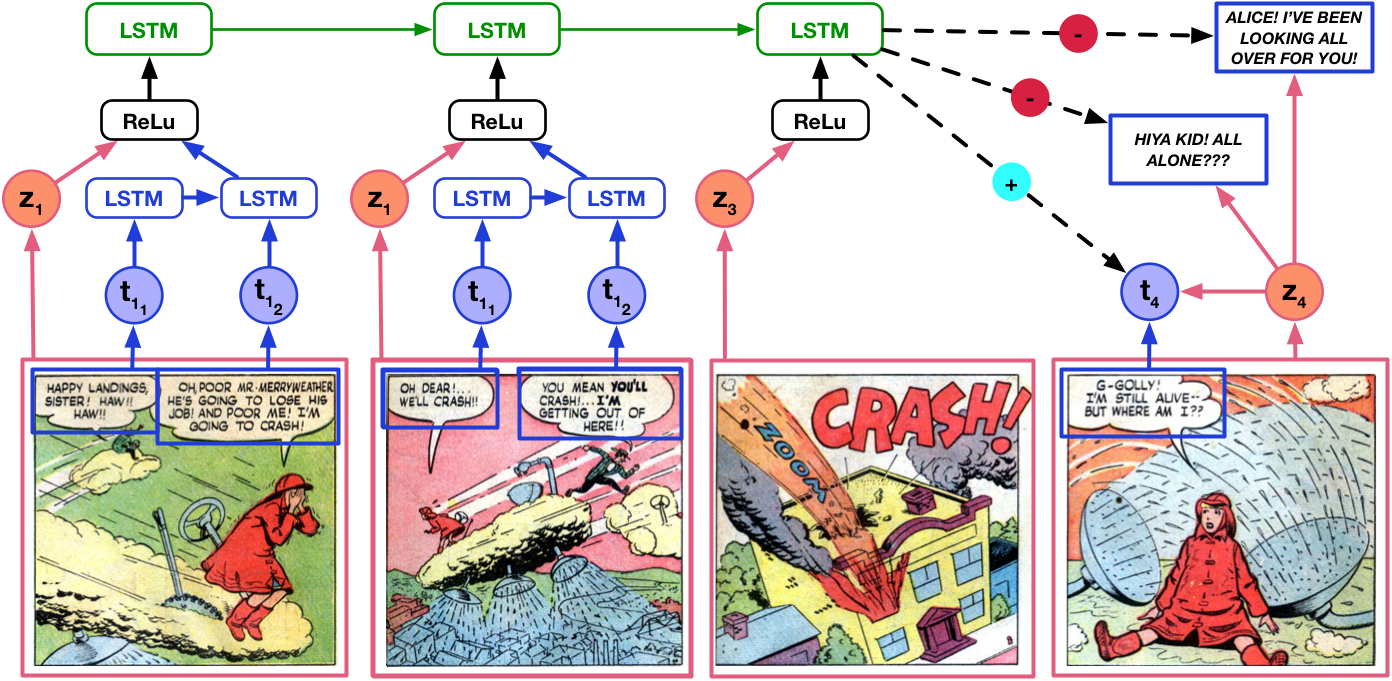}
	\caption{The \textbf{image-text} architecture applied to an instance of the \emph{text cloze} task. Pretrained image features are combined with learned text features in a hierarchical \textsc{lstm} architecture to form a context representation, which is then used to score text candidates.}
	\label{fig:text_cloze}
\end{figure*}

To measure the difficulty of these tasks for deep learning models, we adapt strong baselines for multimodal language and vision understanding tasks to the comics domain. We 
evaluate four different neural models, variants of which were also used to
benchmark the Visual Question Answering dataset~\cite{antol2015vqa} and encode context for visual storytelling~\cite{Huang2016naacl}:
\emph{text-only}, \emph{image-only}, and two \emph{image-text} models. Our best-performing model encodes panels with a hierarchical \textsc{lstm} architecture (see Figure~\ref{fig:text_cloze}). 

On \emph{text cloze}, accuracy
increases when models are given images (in the form of pretrained
\abr{vgg}-16 features) in addition to text; on the other tasks, incorporating both modalities is less important. Additionally, for the \emph{text
  cloze} and \emph{visual cloze} tasks, models perform far worse on
the \emph{hard} setting than the \emph{easy} setting, confirming our
intuition that these tasks are non-trivial when we control for
stylistic dissimilarities between candidates. Finally, none of the
architectures outperform human baselines, which demonstrates the difficulty of understanding \comics: image features obtained from models trained on natural images cannot capture the vast variation in artistic styles, and textual models struggle with the richness and ambiguity of colloquial dialogue highly dependent on visual contexts. In the rest of this section, we first
introduce a shared notation and then use it to specify all of our
models.

\subsection{Model definitions}

In all of our tasks, we are asked to make a prediction about a
particular panel given the preceding $n$ panels as context.\footnote{Test and validation instances for all tasks come from comic books that are unseen during training.}  Each panel
consists of three distinct elements: image, text (\abr{ocr} output), and
textbox bounding box coordinates. For any panel $p_i$, the
corresponding image is $z_i$. Since there can be multiple textboxes
per panel, we refer to individual
textbox contents and bounding boxes as $t_{i_x}$ and $b_{i_x}$, respectively. Each of our tasks
has a different set of answer candidates $A$: \emph{text cloze} has
three text candidates $t_{a_{1\dots3}}$, \emph{visual cloze} has three
image candidates $z_{a_{1\dots3}}$, and \emph{character coherence} has
two combinations of text / bounding box pairs, \{$t_{a_1} / b_{a_1},
t_{a_2} / b_{a_2}$\} and \{$t_{a_1} / b_{a_2}, t_{a_2} /
b_{a_1}$\}. Our architectures differ mainly in the encoding function~$g$ that converts a sequence of context panels $p_{i-1},
p_{i-2}, \dots, p_{i-n}$ into a fixed-length vector $c$. We score the
answer candidates by taking their inner product with $c$ and
normalizing with the softmax function,

\begin{equation}
\label{eq:objective}
s = \text{softmax}(A^{T}c),
\end{equation}

\noindent and we minimize the cross-entropy loss against the ground-truth
labels.\footnote{Performance falters slightly on a development set
  with contrastive max-margin loss functions~\cite{sochergrounded} in
  place of our softmax alternative.}

\paragraph{Text-only:} 
The text-only baseline only has access to the
text $t_{i_x}$ within each panel. Our $g$ function encodes this
text on multiple levels: we first compute a
representation for each $t_{i_x}$ with a word embedding sum\footnote{As in previous work for visual question
  answering~\cite{zhou2015simple}, we
  observe no noticeable improvement with more sophisticated encoding
  architectures.} and then combine multiple textboxes
within the same panel using an
\emph{intrapanel} \textsc{lstm}~\cite{hochreiter1997long}. Finally, we
feed the panel-level representations to an \emph{interpanel}
\textsc{lstm} and take its final hidden state as the context
representation (Figure~\ref{fig:text_cloze}). 
For \emph{text cloze}, the answer candidates are
also encoded with a word embedding sum; for \emph{visual cloze}, we
project the 4096-d \texttt{fc7} layer of VGG-16 down to the word
embedding dimensionality with a fully-connected layer.\footnote{For
  training and testing, we use three panels of context and three candidates. We use a vocabulary size of 30,000 words, restrict the maximum number of textboxes per panel to three, and set the dimensionality of word embeddings and
  \textsc{lstm} hidden states to 256. Models are optimized using
  Adam~\cite{kingma2014adam} for ten epochs, after which we select the
  best-performing model on the dev set.}

\paragraph{Image-only:} 
The image-only baseline is even simpler: we feed the \texttt{fc7} features of
each context panel to an \textsc{lstm} and use the same objective
function as before to score candidates. For \emph{visual cloze}, we
project both the context and answer representations to 512-d with
additional fully-connected layers before scoring. While the
\comics\ dataset is certainly large, we do not attempt learning visual
features from scratch as our task-specific signals are far more
complicated than simple image classification. We also try fine-tuning
the lower-level layers of \abr{vgg}-16~\cite{ayatar2016crossmodal}; however,
this substantially lowers task accuracy even with very small learning
rates for the fine-tuned layers.

\paragraph{Image-text:} 
We combine the previous two models by concatenating the output of the
intrapanel \textsc{lstm} with the \texttt{fc7} representation of the
image and passing the result through a fully-connected layer before
feeding it to the interpanel \textsc{lstm}
(Figure~\ref{fig:text_cloze}). For \emph{text cloze} and
\emph{character coherence}, we also experiment with a variant of the
image-text baseline that has no access to the context panels, which
we dub \textbf{NC-image-text}. In this model, the scoring function
computes inner products between the image features of $p_i$ and the
text candidates.\footnote{We cannot apply this model to \emph{visual cloze} because we are not allowed access to the artwork in panel $p_i$.}
\section{Error Analysis}
\label{sec:discussion}

\begin{table}[t]
\small
\begin{center}
\begin{tabular}{cccccc}
\toprule
\textbf{Model} &  \multicolumn{2}{c}{\footnotesize{\textbf{Text Cloze}}} & \multicolumn{2}{c}{\footnotesize{\textbf{Visual Cloze}}} & \footnotesize{\textbf{Char. Coheren.}} \\
\cmidrule(lr){2-3}\cmidrule(lr){4-5}
& \emph{easy} & \emph{hard} & \emph{easy} & \emph{hard} & \\
\midrule
Random & 33.3 & 33.3 & 33.3 & 33.3 & 50.0 \\
Text-only & 63.4 & 52.9 & 55.9 & 48.4 & 68.2 \\
Image-only & 51.7 & 49.4 & \bf 85.7 & \bf 63.2 & \bf 70.9 \\
NC-image-text & 63.1 & 59.6 & - & - & 65.2 \\
Image-text & \bf 68.6 & \bf 61.0 & 81.3 & 59.1 & 69.3 \\
\midrule
Human & -- & 84 & -- & 88 & 87 \\
\bottomrule
\end{tabular}
\end{center}
\caption{Combining image and text in neural architectures improves their
  ability to predict the next image or dialogue in
  \comics\ narratives. The contextual information present in preceding panels is useful for all tasks: the model that only looks at a single panel (\textbf{NC-image-text}) always underperforms its context-aware counterpart. However, even the best performing models lag well behind humans. }

\label{table:experiments}
\end{table}

Table~\ref{table:experiments} contains our full experimental results,
which we briefly summarize here. On \emph{text cloze}, the image-text model dominates those trained
on a single modality. However, text is much less helpful for \emph{visual cloze} than it is for \emph{text cloze}, suggesting that visual similarity dominates the former task. Having the context of the preceding panels helps
across the board, although the improvements are lower in the
\emph{hard} setting. There is more variation across the
models in the \emph{easy} setting; we hypothesize that the \emph{hard}
case requires moving away from pretrained image features, and transfer
learning methods may prove effective here. Differences between models on \emph{character coherence} are minor; we suspect that more complicated attentional architectures that leverage the bounding box locations $b_{i_x}$ are necessary to ``follow'' speech bubble tails to the characters who speak them.

We also compare all models to a human baseline, for which the authors
manually solve one hundred instances of each task (in the \emph{hard}
setting) given the same preprocessed input that is fed to the neural architectures. Most human errors are the result of poor \abr{ocr} quality (e.g., misspelled words) or low image
resolution. Humans comfortably outperform all models, making it
worthwhile to look at where computers fail but humans succeed.

\begin{figure}[t]
  \centering
    \includegraphics[width=0.9\linewidth]
                   {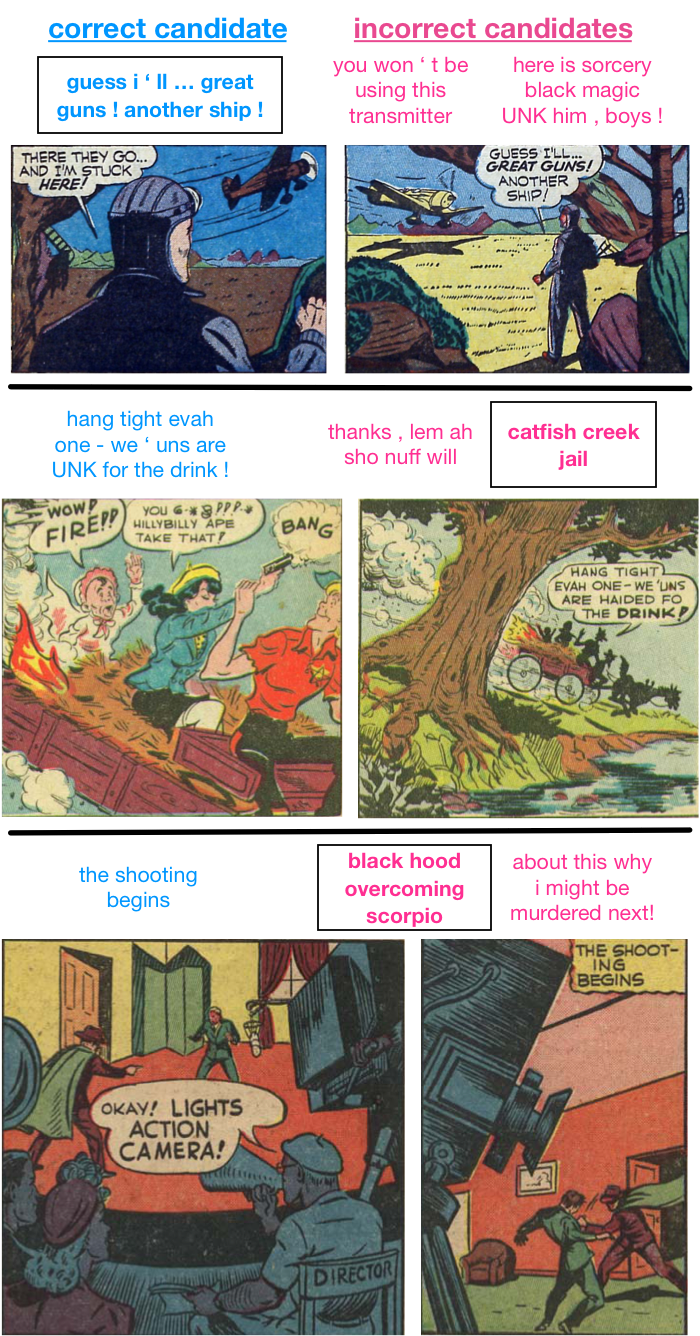}
    \caption{Three \emph{text cloze} examples from the development set, shown with a single panel of context (boxed candidates are predictions by the \textbf{text-image} model). The airplane artwork in the top row helps the \textbf{image-text} model choose the correct answer, while the \textbf{text-only} model fails because the dialogue lacks contextual information. Conversely, the bottom two rows show the
      \textbf{image-text} model ignoring the context in favor of choosing a
      candidate that mentions something visually present in the last
      panel. }  
    	\label{fig:discussion}
\end{figure}

The top row in Figure~\ref{fig:discussion} demonstrates an instance (from \emph{easy text cloze} where the image helps the model make the correct prediction. The text-only model has no idea that an airplane (referred to here as a ``ship'') is present in the panel sequence, as the dialogue in the context panels make no mention of it. In contrast, the image-text model is able to use the artwork to rule out the two incorrect candidates.

The bottom two rows in Figure~\ref{fig:discussion} show \emph{hard text cloze} instances in which
the image-text model is deceived by the artwork in the final panel. While the final panel of the middle row does contain what looks to be a creek, ``catfish creek jail'' is more
suited for a narrative box than a speech bubble, while the meaning of the correct candidate is obscured by the dialect and out-of-vocabulary token.
Similarly, a
camera films a fight scene in the last row; the model selects a candidate that describes a fight instead of focusing on the context in which the scene occurs. 
 These examples
suggest that the contextual information is overridden by strong
associations between text and image, motivating architectures that go
beyond similarity by leveraging external world knowledge to determine
whether an utterance is truly appropriate in a given situation.

\section{Related Work}

Our work is related to three main areas: (1) multimodal tasks that
require language and vision understanding, (2) computational methods
that focus on non-natural images, and (3) models that characterize
language-based narratives.

Deep learning has renewed interest in jointly reasoning about
vision and language. Datasets such as \abr{ms coco}~\cite{mscoco} and
Visual Genome~\cite{krishnavisualgenome} have enabled image
captioning~\cite{vinyals2015show,Karpathycvpr2015,Xu2015show} and
visual question answering~\cite{malinowski2015ask,Lu2016Hie}. Similar to our \emph{character coherence} task, researchers have built models that match \abr{tv} show
characters with their visual attributes~\cite{everinghambuffy} and
speech patterns~\cite{tapaswicharacters}.

Closest to our own comic book setting is the visual storytelling
task, in which systems must generate~\cite{huang2016visual} or
reorder~\cite{AgrawalCBPB16} stories given a dataset (\textsc{sind}) of photos from Flikr
galleries of ``storyable'' events such as weddings and birthday
parties. \textsc{sind}'s images are fundamentally different from
\comics\ in that they lack coherent characters and accompanying
dialogue. Comics are created by skilled professionals, not
crowdsourced workers, and they offer a far greater variety of
character-centric stories that depend on dialogue to further the
narrative; with that said, the text in \comics\ is less suited for generation because of \abr{ocr}
errors.

We build here on previous work that attempts to understand non-natural
images. Zitnick et al.~\cite{ZitnickAbstract2016} discover semantic
scene properties from a clip art dataset featuring characters and
objects in a limited variety of settings. Applications of deep learning to
paintings include tasks such as detecting objects in oil
paintings~\cite{crowley2014state,Crowley15} and answering
questions about
artwork~\cite{Guha:Iyyer:Boyd-Graber-2016}. Previous computational
work on comics focuses primarily on extracting elements such as panels
and textboxes~\cite{RigaudPhD}; in addition to the references in
Section~\ref{sec:dataset}, there is a large body of segmentation
research on
manga~\cite{AramkiMangaSegmentation,Pang14MangaSegmentation,Matsui15,KovanenMangaReadingOrder}.

To the best of our knowledge, we are the first to computationally
model \emph{content} in comic books as opposed to just extracting
their elements. We follow previous work in language-based narrative
understanding; very similar to our \emph{text cloze} task is the
``Story Cloze Test''~\cite{mostafazadeh-EtAl:2016:N16-1}, in which
models must predict the ending to a short (four sentences long)
story. Just like our tasks, the Story Cloze Test proves difficult for
computers and motivates future research into commonsense knowledge
acquisition. Others have studied
characters~\cite{elson2010extracting,bamman-underwood-smith:2014:P14-1,IyyerRelationships2016}
and narrative
structure~\cite{schankabelson77,lehnert1981plot,chambers2009unsupervised}
in novels.
\section{Conclusion \& Future Work}
\label{sec:conclusion}

We present the \comics\ dataset, which contains over 1.2 million
panels from ``Golden Age'' comic books. We design three cloze-style
tasks on \comics\ to explore \emph{closure}, or how readers connect
disparate panels into coherent stories. Experiments with different
neural architectures, along with a manual data analysis, confirm the importance of multimodal models that combine text and image for comics understanding. We additionally show that context is crucial for
predicting narrative or character-centric aspects of panels.

However, for computers to reach human performance, they will need to
become better at leveraging context. Readers rely on commonsense
knowledge to make sense of dramatic scene and camera changes; how can
we inject such knowledge into our models? Another potentially
intriguing direction, especially given recent advances in generative
adversarial networks~\cite{goodfellow2014generative}, is generating
artwork given dialogue (or vice versa). Finally, \comics\ presents a
golden opportunity for transfer learning; can we train models that
generalize across natural and non-natural images much like humans do?

\section{Acknowledgments}
\label{sec:acknowledgments}

We thank the anonymous reviewers for their insightful comments and the UMIACS and Google Cloud Support staff for their help with \abr{ocr}. Manjunatha and Davis were supported by Office of Naval Research grant N000141612713, while Iyyer, Boyd-Graber, and Daum\'e were supported by NSF grant IIS-1320538. Any opinions, findings, or conclusions expressed here are those of the authors and do not necessarily reflect the view of the sponsor.

\clearpage

{\small
\bibliographystyle{ieee}
\bibliography{journal-full,comics}

\begin{thebibliography}{10}\itemsep=-1pt

\bibitem{AgrawalCBPB16}
H.~Agrawal, A.~Chandrasekaran, D.~Batra, D.~Parikh, and M.~Bansal.
\newblock Sort story: Sorting jumbled images and captions into stories.
\newblock In {\em Proceedings of Empirical Methods in Natural Language
  Processing}, 2016.

\bibitem{antol2015vqa}
S.~Antol, A.~Agrawal, J.~Lu, M.~Mitchell, D.~Batra, C.~Lawrence~Zitnick, and
  D.~Parikh.
\newblock Vqa: Visual question answering.
\newblock In {\em International Conference on Computer Vision}, 2015.

\bibitem{AramkiMangaSegmentation}
Y.~Aramaki, Y.~Matsui, T.~Yamasaki, and K.~Aizawa.
\newblock Interactive segmentation for manga.
\newblock In {\em Special Interest Group on Computer Graphics and Interactive
  Techniques Conference}, 2014.

\bibitem{ayatar2016crossmodal}
Y.~Aytar, L.~Castrejon, C.~Vondrick, H.~Pirsiavash, and A.~Torralba.
\newblock Cross-modal scene networks.
\newblock {\em arXiv}, 2016.

\bibitem{bamman-underwood-smith:2014:P14-1}
D.~Bamman, T.~Underwood, and N.~A. Smith.
\newblock A {B}ayesian mixed effects model of literary character.
\newblock In {\em Proceedings of the Association for Computational
  Linguistics}, 2014.

\bibitem{berg2013unsupervised}
T.~Berg-Kirkpatrick, G.~Durrett, and D.~Klein.
\newblock Unsupervised transcription of historical documents.
\newblock In {\em Proceedings of the Association for Computational
  Linguistics}, 2013.

\bibitem{carroll1992visual}
P.~J. Carroll, J.~R. Young, and M.~S. Guertin.
\newblock Visual analysis of cartoons: A view from the far side.
\newblock In {\em Eye movements and visual cognition}. Springer, 1992.

\bibitem{chambers2009unsupervised}
N.~Chambers and D.~Jurafsky.
\newblock Unsupervised learning of narrative schemas and their participants.
\newblock In {\em Proceedings of the Association for Computational
  Linguistics}, 2009.

\bibitem{chatfield2014return}
K.~Chatfield, K.~Simonyan, A.~Vedaldi, and A.~Zisserman.
\newblock Return of the devil in the details: Delving deep into convolutional
  nets.
\newblock In {\em British Machine Vision Conference}, 2014.

\bibitem{cohn2010limits}
N.~Cohn.
\newblock The limits of time and transitions: challenges to theories of
  sequential image comprehension.
\newblock {\em Studies in Comics}, 1(1), 2010.

\bibitem{crowley2014state}
E.~Crowley and A.~Zisserman.
\newblock The state of the art: Object retrieval in paintings using
  discriminative regions.
\newblock In {\em British Machine Vision Conference}, 2014.

\bibitem{Crowley15}
E.~J. Crowley, O.~M. Parkhi, and A.~Zisserman.
\newblock Face painting: querying art with photos.
\newblock In {\em British Machine Vision Conference}, 2015.

\bibitem{eisner1990comics}
W.~Eisner.
\newblock {\em Comics \& Sequential Art}.
\newblock Poorhouse Press, 1990.

\bibitem{elson2010extracting}
D.~K. Elson, N.~Dames, and K.~R. McKeown.
\newblock Extracting social networks from literary fiction.
\newblock In {\em Proceedings of the Association for Computational
  Linguistics}, 2010.

\bibitem{everinghambuffy}
M.~Everingham, J.~Sivic, and A.~Zisserman.
\newblock Hello! my name is... {B}uffy'' -- automatic naming of characters in
  {TV} video.
\newblock In {\em Proceedings of the British Machine Vision Conference}, 2006.

\bibitem{cohn16}
T.~Foulsham, D.~Wybrow, and N.~Cohn.
\newblock Reading without words: Eye movements in the comprehension of comic
  strips.
\newblock {\em Applied Cognitive Psychology}, 30, 2016.

\bibitem{goodfellow2014generative}
I.~Goodfellow, J.~Pouget-Abadie, M.~Mirza, B.~Xu, D.~Warde-Farley, S.~Ozair,
  A.~Courville, and Y.~Bengio.
\newblock Generative adversarial nets.
\newblock In {\em Proceedings of Advances in Neural Information Processing
  Systems}, 2014.

\bibitem{goulart2004comic}
R.~Goulart.
\newblock {\em Comic Book Encyclopedia: The Ultimate Guide to Characters,
  Graphic Novels, Writers, and Artists in the Comic Book Universe}.
\newblock HarperCollins, 2004.

\bibitem{guerin2013ebdtheque}
C.~Gu{\'e}rin, C.~Rigaud, A.~Mercier, F.~Ammar-Boudjelal, K.~Bertet, A.~Bouju,
  J.-C. Burie, G.~Louis, J.-M. Ogier, and A.~Revel.
\newblock e{BD}theque: a representative database of comics.
\newblock In {\em International Conference on Document Analysis and
  Recognition}, 2013.

\bibitem{Guha:Iyyer:Boyd-Graber-2016}
A.~Guha, M.~Iyyer, and J.~Boyd-Graber.
\newblock A distorted skull lies in the bottom center: Identifying paintings
  from text descriptions.
\newblock In {\em NAACL Human-Computer Question Answering Workshop}, 2016.

\bibitem{tapaswicharacters}
M.~Haurilet, M.~Tapaswi, Z.~Al{-}Halah, and R.~Stiefelhagen.
\newblock Naming {TV} characters by watching and analyzing dialogs.
\newblock In {\em {IEEE} Winter Conference on Applications of Computer Vision},
  2016.

\bibitem{ho2012panel}
A.~K.~N. Ho, J.-C. Burie, and J.-M. Ogier.
\newblock Panel and speech balloon extraction from comic books.
\newblock In {\em IAPR International Workshop on Document Analysis Systems},
  2012.

\bibitem{hochreiter1997long}
S.~Hochreiter and J.~Schmidhuber.
\newblock Long short-term memory.
\newblock {\em Neural computation}, 1997.

\bibitem{huang2016visual}
T.-H.~K. Huang, F.~Ferraro, N.~Mostafazadeh, I.~Misra, A.~Agrawal, J.~Devlin,
  R.~Girshick, X.~He, P.~Kohli, D.~Batra, et~al.
\newblock Visual storytelling.
\newblock In {\em Conference of the North American Chapter of the Association
  for Computational Linguistics}, 2016.

\bibitem{Huang2016naacl}
T.~K. Huang, F.~Ferraro, N.~Mostafazadeh, I.~Misra, A.~Agrawal, J.~Devlin,
  R.~B. Girshick, X.~He, P.~Kohli, D.~Batra, C.~L. Zitnick, D.~Parikh,
  L.~Vanderwende, M.~Galley, and M.~Mitchell.
\newblock Visual storytelling.
\newblock In {\em Conference of the North American Chapter of the Association
  for Computational Linguistics}, 2016.

\bibitem{IyyerRelationships2016}
M.~Iyyer, A.~Guha, S.~Chaturvedi, J.~Boyd-Graber, and H.~{Daum\'{e} III}.
\newblock Feuding families and former friends: Unsupervised learning for
  dynamic fictional relationships.
\newblock In {\em Conference of the North American Chapter of the Association
  for Computational Linguistics}, 2016.

\bibitem{jaderberg2016reading}
M.~Jaderberg, K.~Simonyan, A.~Vedaldi, and A.~Zisserman.
\newblock Reading text in the wild with convolutional neural networks.
\newblock {\em International Journal of Computer Vision}, 116(1), 2016.

\bibitem{Karpathycvpr2015}
A.~Karpathy and F.~Li.
\newblock Deep visual-semantic alignments for generating image descriptions.
\newblock In {\em {IEEE} Conference on Computer Vision and Pattern Recognition,
  {CVPR} 2015, Boston, MA, USA, June 7-12, 2015}, 2015.

\bibitem{kingma2014adam}
D.~Kingma and J.~Ba.
\newblock Adam: A method for stochastic optimization.
\newblock In {\em Proceedings of the International Conference on Learning
  Representations}, 2014.

\bibitem{KovanenMangaReadingOrder}
S.~Kovanen and K.~Aizawa.
\newblock A layered method for determining manga text bubble reading order.
\newblock In {\em International Conference on Image Processing}, 2015.

\bibitem{krishnavisualgenome}
R.~Krishna, Y.~Zhu, O.~Groth, J.~Johnson, K.~Hata, J.~Kravitz, S.~Chen,
  Y.~Kalantidis, L.-J. Li, D.~A. Shamma, M.~Bernstein, and L.~Fei-Fei.
\newblock Visual genome: Connecting language and vision using crowdsourced
  dense image annotations.
\newblock 2016.

\bibitem{krizhevsky2012imagenet}
A.~Krizhevsky, I.~Sutskever, and G.~E. Hinton.
\newblock Imagenet classification with deep convolutional neural networks.
\newblock In {\em Proceedings of Advances in Neural Information Processing
  Systems}, 2012.

\bibitem{lehnert1981plot}
W.~G. Lehnert.
\newblock Plot units and narrative summarization.
\newblock {\em Cognitive Science}, 5(4), 1981.

\bibitem{li2014automatic}
L.~Li, Y.~Wang, Z.~Tang, and L.~Gao.
\newblock Automatic comic page segmentation based on polygon detection.
\newblock {\em Multimedia Tools and Applications}, 69(1), 2014.

\bibitem{mscoco}
T.~Lin, M.~Maire, S.~J. Belongie, L.~D. Bourdev, R.~B. Girshick, J.~Hays,
  P.~Perona, D.~Ramanan, P.~Doll{\'{a}}r, and C.~L. Zitnick.
\newblock Microsoft {COCO:} common objects in context.
\newblock 2014.

\bibitem{Lu2016Hie}
J.~Lu, J.~Yang, D.~Batra, and D.~Parikh.
\newblock Hierarchical question-image co-attention for visual question
  answering, 2016.

\bibitem{malinowski2015ask}
M.~Malinowski, M.~Rohrbach, and M.~Fritz.
\newblock Ask your neurons: A neural-based approach to answering questions
  about images.
\newblock In {\em Computer Vision and Pattern Recognition}, 2015.

\bibitem{Matsui15}
Y.~Matsui.
\newblock Challenge for manga processing: Sketch-based manga retrieval.
\newblock In {\em Proceedings of the 23rd Annual {ACM} Conference on
  Multimedia}, 2015.

\bibitem{matsui2015sketch}
Y.~Matsui, K.~Ito, Y.~Aramaki, T.~Yamasaki, and K.~Aizawa.
\newblock Sketch-based manga retrieval using manga109 dataset.
\newblock {\em arXiv preprint arXiv:1510.04389}, 2015.

\bibitem{mccloud1994understanding}
S.~McCloud.
\newblock {\em Understanding Comics}.
\newblock HarperCollins, 1994.

\bibitem{morton1966computer}
G.~M. Morton.
\newblock {\em A computer oriented geodetic data base and a new technique in
  file sequencing}.
\newblock International Business Machines Co, 1966.

\bibitem{mostafazadeh-EtAl:2016:N16-1}
N.~Mostafazadeh, N.~Chambers, X.~He, D.~Parikh, D.~Batra, L.~Vanderwende,
  P.~Kohli, and J.~Allen.
\newblock A corpus and cloze evaluation for deeper understanding of commonsense
  stories.
\newblock In {\em Conference of the North American Chapter of the Association
  for Computational Linguistics}, 2016.

\bibitem{pang2014robust}
X.~Pang, Y.~Cao, R.~W. Lau, and A.~B. Chan.
\newblock A robust panel extraction method for manga.
\newblock In {\em Proceedings of the {ACM} International Conference on
  Multimedia}, 2014.

\bibitem{Pang14MangaSegmentation}
X.~Pang, Y.~Cao, R.~W.~H. Lau, and A.~B. Chan.
\newblock A robust panel extraction method for manga.
\newblock In {\em Proceedings of the {ACM} International Conference on
  Multimedia}, 2014.

\bibitem{renNIPS15fasterrcnn}
S.~Ren, K.~He, R.~Girshick, and J.~Sun.
\newblock Faster {R-CNN}: Towards real-time object detection with region
  proposal networks.
\newblock In {\em Proceedings of Advances in Neural Information Processing
  Systems}, 2015.

\bibitem{RigaudPhD}
C.~Rigaud.
\newblock {\em Segmentation and indexation of complex objects in comic book
  images}.
\newblock PhD thesis, University of La Rochelle, France, 2014.

\bibitem{rigaud2013active}
C.~Rigaud, J.-C. Burie, J.-M. Ogier, D.~Karatzas, and J.~Van~de Weijer.
\newblock An active contour model for speech balloon detection in comics.
\newblock In {\em International Conference on Document Analysis and
  Recognition}, 2013.

\bibitem{rigaud2015knowledge}
C.~Rigaud, C.~Gu{\'e}rin, D.~Karatzas, J.-C. Burie, and J.-M. Ogier.
\newblock Knowledge-driven understanding of images in comic books.
\newblock {\em International Journal on Document Analysis and Recognition},
  18(3), 2015.

\bibitem{schankabelson77}
R.~Schank and R.~Abelson.
\newblock {\em Scripts, Plans, Goals and Understanding: an Inquiry into Human
  Knowledge Structures}.
\newblock L. Erlbaum, 1977.

\bibitem{smith2007overview}
R.~Smith.
\newblock An overview of the tesseract ocr engine.
\newblock In {\em International Conference on Document Analysis and
  Recognition}, 2007.

\bibitem{sochergrounded}
R.~Socher, Q.~V. Le, C.~D. Manning, and A.~Y. Ng.
\newblock Grounded compositional semantics for finding and describing images
  with sentences.
\newblock {\em Transactions of the Association for Computational Linguistics},
  2014.

\bibitem{tanaka2007layout}
T.~Tanaka, K.~Shoji, F.~Toyama, and J.~Miyamichi.
\newblock Layout analysis of tree-structured scene frames in comic images.
\newblock In {\em International Joint Conference on Artificial Intelligence},
  2007.

\bibitem{taylor1953cloze}
W.~L. Taylor.
\newblock Cloze procedure: a new tool for measuring readability.
\newblock {\em Journalism and Mass Communication Quarterly}, 30(4), 1953.

\bibitem{vinyals2015show}
O.~Vinyals, A.~Toshev, S.~Bengio, and D.~Erhan.
\newblock Show and tell: A neural image caption generator.
\newblock In {\em Computer Vision and Pattern Recognition}, 2015.

\bibitem{xiong2016dynamic}
C.~Xiong, S.~Merity, and R.~Socher.
\newblock Dynamic memory networks for visual and textual question answering.
\newblock In {\em Proceedings of the International Conference of Machine
  Learning}, 2016.

\bibitem{Xu2015show}
K.~Xu, J.~Ba, R.~Kiros, K.~Cho, A.~Courville, R.~Salakhutdinov, R.~Zemel, and
  Y.~Bengio.
\newblock Show, attend and tell: Neural image caption generation with visual
  attention.
\newblock In {\em Proceedings of the International Conference of Machine
  Learning}, 2015.

\bibitem{zhou2015simple}
B.~Zhou, Y.~Tian, S.~Sukhbaatar, A.~Szlam, and R.~Fergus.
\newblock Simple baseline for visual question answering.
\newblock {\em arXiv preprint arXiv:1512.02167}, 2015.

\bibitem{ZitnickAbstract2016}
C.~L. Zitnick, R.~Vedantam, and D.~Parikh.
\newblock Adopting abstract images for semantic scene understanding.
\newblock {\em {IEEE} Trans. Pattern Anal. Mach. Intell.}, 38(4):627--638,
  2016.

\end{thebibliography}
}

\end{document}


\maketitle

\section{\abr{ocr} Post-Processing and Advertisement Removal}
\abr{ocr} makes systematic mistakes on
our textboxes. We target two types of these mistakes using
PyEnchant:\footnote{\url{http://pythonhosted.org/pyenchant/faq.html}}
1) where the \abr{ocr} system fails to recognize the first letter of a
particular word (e.g., \emph{eleportation} instead of
\emph{teleportation}), and 2) where the \abr{ocr} system transcribes part of a word as a
single alphabetical character. To eliminate errors of the first type,
we start by tokenizing the \abr{ocr} output using NLTK's Punkt Tokenizer.\footnote{\url{http://www.nltk.org/}} We then sort the vocabulary of the tokenized \abr{ocr}
output in decreasing order of frequency and pick words ranked from
10,001 to 100,000, because most misspelled words are also rare. For each of these words that is length three or
longer, we look up the most likely suggestion offered by PyEnchant. If
the only difference between the most likely suggestion and the
original word is an additional letter in the first position of the
suggestion, then we replace the word with the suggestion everywhere in
our corpus. To correct the second type of errors, we simply delete all
single character alphabetical tokens that are not one of 'a', 'd',
'i', 'm', 's', 't' - characters which can plausibly occur by
themselves quite frequently (some occur after an apostrophe).

In addition to spelling errors, the books in \textsc{comics} contain many advertisements that we need to remove before generating data for our tasks.
While most dialogue and narration boxes contain less than 30 words,
longer textboxes frequently come from full-page product
advertisements (e.g., Figure~\ref{fig:ad}). However, detecting ads from page images is not easy. Some ads are deceptively similar to comic pages,
containing images and even containing faux mini-comics. Aside from ads, there are also other undesirable pages; many books contain text-only short stories in addition to comics. We remove these kinds of pages using features from \abr{ocr} transcriptions. We annotate each page of 100 random books with a label
indicating the presence or absence of an invalid page as our training
set and each page of 20 random books as our test set. Out of 6,117 annotated pages, 697 of them are either advertisements or text-only stories (11.4\%). We train a binary classifier using Vowpal Wabbit:\footnote{\url{https://github.com/JohnLangford/vowpal_wabbit/wiki}} which takes the \abr{ocr} text for all the panels of a pages as lexical
features (unigrams and bigrams). We improve our model by adding features like total
count of words in the page and a count of non-alphanumeric characters. Our model gives us
a total misclassification error of 8\% and a false negative error of
17.3\%, which means it misses one invalid page out of every six. The
model has a negligible false positive error of 0.2\%. Using this model
to filter the entire dataset of 198,657 pages yields 13,200 invalid
pages.

\begin{figure*}
        \includegraphics[width=\linewidth, height=22cm]
        {figures/ad.jpg}
        \caption{An advertisement from the dataset. The juxtaposition of text and image causes it to slightly resemble a comics page.}
        \label{fig:ad}
\end{figure*}

\section{Examples from Dataset Creation}
\abr{ocr} transcription is the final stage of our data creation pipeline (panel extraction $\rightarrow$ textbox extraction $\rightarrow$ \abr{ocr}). Therefore, faulty outputs in any of the preceeding steps can lead to faulty \abr{ocr} outputs. In Figure~\ref{fig:OCR1}, there are only minor errors in \abr{ocr} extraction due to understandable misinterpretations of the text in the dialog boxes. For example, the \abr{ocr} interprets the letters ``IC" as ``K", which leads to incorrectly predicting the word ``QUICKLY" as ``QUKKLY". However, in Figure~\ref{fig:OCR2}, we observe a more critical error due to missing pixels in the panel extraction process. Due to the layout of the textbox in the panel, crucial portions of the text are trimmed from view; while the \abr{ocr} does a valiant job of predicting the contents of the textbox, its output is gibberish.
\begin{figure*}
        \centering
        \includegraphics[height=6cm]
        {figures/OCR_fig1.png}
        \caption{A minor \abr{ocr} error. Mistakes such as predicting ``BG" for ``BIG'' are understandable, since the `I' in ``BIG" is barely visible. Similarly, the ``IC" in ``QUICKLY'' looks a lot like ``K'' in this font. Finally, ``SUB STANCE" is predicted rather than ``SUBSTANCE", due to an end-of-line word break.}
        \label{fig:OCR1}
\end{figure*}

\begin{figure*}
        \includegraphics[width=\linewidth]
        {figures/OCR_fig2.png}
        \caption{A major \abr{ocr} error. In part a) of the figure, note the location of the panel in the page. b) gives us the panel as predicted by the RCNN, but a critical portion of the text is missing. As a consequence, the textbox extraction is also faulty, rendering the \abr{ocr} completely meaningless.}
        \label{fig:OCR2}
\end{figure*}